\documentclass[sigconf,natbib=true]{acmart}

\copyrightyear{2026}
\acmYear{2026}
\setcopyright{cc}
\setcctype{by}
\acmConference[CIKM '26]{Proceedings of the 35th ACM International Conference on Information and Knowledge Management}{November 07--11, 2026}{Rome, Italy}
\acmBooktitle{Proceedings of the 35th ACM International Conference on Information and Knowledge Management (CIKM '26), November 07--11, 2026, Rome, Italy}
\acmDOI{10.1145/3799682.3840026}
\acmISBN{979-8-4007-2539-5/2026/11}

\usepackage{booktabs}
\usepackage{multirow}
\usepackage{xcolor}
\usepackage{subcaption}
\usepackage{xspace}
\usepackage{algorithm}
\usepackage{algorithmic}
\usepackage[capitalize]{cleveref}

\definecolor{ourgreen}{HTML}{2E7D32}
\newcommand{\method}{AdaMerge\xspace}

\begin{document}

\title{\method: Tuning-Free Patch Compression for Multi-Vector Visual Document Retrieval}

\author{Jianxin You}
\affiliation{%
  \institution{Universit\'e de Montr\'eal}
  \city{Montr\'eal}
  \state{QC}
  \country{Canada}}
\email{jianxin.you@umontreal.ca}

\author{Kun Ni}
\affiliation{%
  \institution{Concordia University}
  \city{Montr\'eal}
  \state{QC}
  \country{Canada}}
\email{kun.ni@mail.concordia.ca}

\begin{abstract}
Multi-vector visual document retrieval (VDR) models such as ColPali and
ColNomic achieve strong accuracy by representing each document with hundreds
to thousands of patch-level embeddings, at substantial storage and latency
cost. Existing compression methods either prune unimportant patches or merge
similar ones into clusters; the recent state-of-the-art merging method
Prune-then-Merge (PtM) consistently outperforms pruning-only baselines at high
compression, but requires a per-dataset cluster budget $m$ to be tuned by grid
search. We
observe that the merge-cosine sequence produced by hierarchical clustering
exhibits a sharp \emph{cliff} separating mergeable redundancy from salient
signal, and that the location of this cliff is concentrated in a narrow band
across more than $11{,}000$ documents from 14 datasets. This
suggests the merge boundary can be detected \emph{per document} rather than
tuned per dataset. Building on this observation, we propose \method, a
plug-and-play compression method that (i)~detects each document's own cliff via
gap analysis on the merge-cosine trajectory, and (ii)~builds attention-weighted
cluster centroids to preserve salient signal. On the long-document benchmark
ViDoRe-V2 (4 datasets, two backbones), \method significantly outperforms tuned
PtM across the operating range ($p<10^{-4}$); on the short-document benchmark
ViDoRe-V1 (10 datasets, two backbones), where all merging methods are already
near-lossless, \method matches tuned PtM without any per-dataset tuning. \method adds only
${\approx}10$~ms per document and exposes a single global hyperparameter shared
across all datasets and backbones.
\end{abstract}



\maketitle

\section{Introduction}
\label{sec:intro}

Visual document retrieval (VDR) treats each document page as an image and
retrieves it directly with a vision--language encoder, sidestepping the
brittle OCR-and-parse pipeline of text-based retrieval. Multi-vector VDR models
in the ColPali family~\cite{faysse2024colpali,nomicai2025} push this further:
rather than a single embedding per page, they emit one embedding per visual
patch (hundreds to thousands per page) and score documents with a
late-interaction \emph{MaxSim} operator~\cite{khattab2020colbert}. This
fine-grained representation is the source of their state-of-the-art accuracy,
but also of their cost: a single page can occupy tens of kilobytes, and an
index of a modest corpus quickly reaches hundreds of megabytes, with retrieval
latency growing in the number of stored patches.

Two families of methods compress these representations. \emph{Pruning} methods
such as DocPruner~\cite{yan2025docpruner} discard low-attention patches;
\emph{merging} methods cluster similar patches and keep one representative per
cluster. The recent Prune-then-Merge (PtM)~\cite{yan2026ptm} combines both and
defines the current state of the art, dominating pruning-only baselines at high
compression. PtM's effectiveness, however, hinges on a discrete cluster budget
$m$ that sets how aggressively each document is merged. The optimal $m$ varies
across corpora, so deploying PtM on a new collection requires a per-dataset grid
search over $m$, an offline cost that is easy to overlook in a benchmark table
but real in practice.

We revisit \emph{where} merging should stop. Running Ward clustering on a
document's patches produces a monotonically decreasing sequence of merge
similarities; we observe that this sequence does not decay smoothly but drops
sharply at a \emph{cliff} that separates near-duplicate patches (safe to merge)
from semantically distinct ones (costly to merge). Across $11{,}273$ documents
spanning 14 datasets, the cliff location is concentrated in a
narrow band rather than scattered: the right stopping point is a property each
document reveals on its own, not a budget to be searched per dataset.

Building on this observation we propose \method, a tuning-free compression
method that detects each document's cliff by locating the largest gap in its
merge-cosine sequence, and forms attention-weighted centroids for the resulting
clusters. \method is plug-and-play: it reuses the same pruning front-end as
PtM, adds only ${\approx}10$~ms per document, and replaces PtM's per-dataset
budget $m$ with a single global constant. Our contributions are:
\begin{itemize}\setlength\itemsep{2pt}
\item We identify a sharp, consistently located \emph{cliff} in the per-document
merge-cosine sequence, and show it can be detected per document instead of tuned
per dataset (\Cref{sec:cliff,sec:exp-cliff}).
\item We propose \method, a tuning-free merging method built on per-document
cliff detection and attention-weighted centroids (\Cref{sec:adamerge}).
\item We show \method significantly beats tuned PtM on long documents
(ViDoRe-V2, two backbones, $p<10^{-4}$) and matches it on short documents
(ViDoRe-V1) with no per-dataset tuning (\Cref{sec:exp}).
\end{itemize}

\section{Related Work}
\label{sec:related}

\textbf{Multi-vector visual document retrieval.}
Late-interaction retrieval originates with ColBERT~\cite{khattab2020colbert},
which represents queries and documents as token-level embeddings and scores them
with MaxSim; PLAID~\cite{santhanam2022plaid} makes this practical at scale.
ColPali~\cite{faysse2024colpali} ports the idea to the visual domain, encoding
page images into patch embeddings with a vision--language model, and
ColNomic~\cite{nomicai2025} improves the backbone. The ViDoRe
benchmarks~\cite{faysse2024colpali,mace2025vidorev2} are the standard evaluation
suites. These models' accuracy comes from retaining per-patch detail, which is
precisely what makes their indexes expensive.

\textbf{Compressing multi-vector representations.}
One line of work prunes: DocPruner~\cite{yan2025docpruner} removes patches with
low end-of-sequence attention. Another merges redundant patches; clustering them
and keeping cluster centroids (which we call Sem-Cluster, the merge-only
baseline studied in~\cite{ma2025light}) reduces count without an explicit
importance signal. Prune-then-Merge~\cite{yan2026ptm} combines pruning and
clustering and is the strongest prior method, but fixes the number of clusters
through a per-dataset budget. Orthogonal efforts compress at the token level for
LLM-based retrieval~\cite{wen2025token,zong2025lossless} or learn compact
representations end-to-end~\cite{xiao2025metaembed}. \method differs
in that it neither fixes a budget nor learns one: it reads the stopping point off
each document's own merge trajectory.

\section{Method}
\label{sec:method}

\subsection{Problem Setup}
A multi-vector VDR model encodes a document $D$ into a set of $N$ patch
embeddings $\mathbf{P}=\{\mathbf{p}_1,\dots,\mathbf{p}_N\}$ with
$\mathbf{p}_i\in\mathbb{R}^{d}$ ($d{=}128$ for ColPali-style backbones). Each
patch carries an EOS-attention score $\alpha_i\in\mathbb{R}_{\ge 0}$. Given a
query $Q$ with embeddings $\{\mathbf{q}_1,\dots,\mathbf{q}_L\}$, documents are
scored by late-interaction MaxSim~\cite{khattab2020colbert}:
\[
\mathrm{Score}(Q,D)=\sum_{i=1}^{L}\max_{j\le N}
\langle \mathbf{q}_i,\mathbf{p}_j\rangle.
\]
Compression produces a compact set
$\widetilde{\mathbf{P}}=\{\widetilde{\mathbf{p}}_1,\dots,
\widetilde{\mathbf{p}}_M\}$ with $M\!\ll\!N$ while preserving retrieval quality.
We report the \emph{compression rate} $1-M/N$.

\subsection{Background: Prune-then-Merge}
\label{sec:bg-ptm}
PtM~\cite{yan2026ptm} operates in two stages. \textbf{Prune}: following
DocPruner~\cite{yan2025docpruner}, it keeps patches whose attention exceeds an
adaptive threshold $\mu_{\alpha}+k\,\sigma_{\alpha}$, where
$\mu_{\alpha},\sigma_{\alpha}$ are the within-document mean and standard
deviation of $\alpha$ and $k$ controls aggressiveness, leaving $N'\!\le\!N$
patches. \textbf{Merge}: it runs Ward hierarchical clustering on the
L2-normalised survivors, stops at $\lceil N'/m\rceil$ clusters, and outputs the
unweighted mean of each cluster. The budget $m\!\in\!\mathbb{Z}_{\ge1}$ is a
discrete hyperparameter whose optimum shifts across datasets, requiring a grid
search per benchmark.

\subsection{The Merge-Cosine Cliff}
\label{sec:cliff}
As Ward merges the two most similar clusters at each step, it produces a
monotonically non-increasing \emph{merge-cosine sequence}
$\mathbf{c}=(c_1,\dots,c_{N'-1})$, where $c_t$ is the cosine between the two
centroids merged at step $t$ (on normalised embeddings this is equivalent to
Ward distance up to a monotone transform). Early merges combine near-duplicate
patches with $c_t\!\approx\!1$; late merges combine semantically distinct
clusters with $c_t\!\to\!0$.

Empirically this sequence does not decay smoothly: it drops sharply at a
\emph{cliff} that marks the boundary between mergeable redundancy and salient
information. Merging past the cliff fuses clusters that are no longer
interchangeable, producing centroids that average across genuinely distinct
content and erase signal. \Cref{sec:exp-cliff} shows the cliff location is
concentrated across $11{,}273$ documents from 14 datasets.

\subsection{\method}
\label{sec:adamerge}
\method shares Stage~1 (DocPruner pruning) with PtM and replaces the
fixed-budget cut and unweighted centroid with two tuning-free components:

\begin{itemize}\setlength\itemsep{2pt}
\item \textbf{Adaptive cut (per document).} Compute the merge-cosine sequence
$\mathbf{c}$. Among consecutive pairs with $c_t\!\ge\!c_{\min}$, locate the
largest drop $t^\star=\arg\max_{t}(c_t-c_{t+1})$ and stop merging there. The
constant $c_{\min}$ is a global \emph{safety floor} set once across all
datasets (not a per-dataset budget); it prevents degenerate documents from
collapsing to a single cluster.
\item \textbf{Attention-weighted centroid.} For each surviving cluster
$\mathcal{C}_k$, output
\[
\widetilde{\mathbf{p}}_k=\frac{\sum_{j\in\mathcal{C}_k}
\alpha_j\,\mathbf{p}_j}{\sum_{j\in\mathcal{C}_k}\alpha_j},
\]
biasing the representative toward high-attention patches, which carry more
query-relevant signal than low-attention members.
\end{itemize}

\Cref{alg:adamerge} summarizes the pipeline. Because the cut is selected per
document from $\mathbf{c}$ alone, \method has no per-dataset tuning step: a
single $c_{\min}$ is used across all benchmarks, and the pruning aggressiveness
$k$ (shared with PtM) selects the operating compression rate. Throughout, \emph{tuning-free} thus means free of \emph{per-dataset} (and
per-backbone) hyperparameter search: $c_{\min}$ is a coarse global constant
fixed once, not a budget to be re-searched on each new corpus.

\begin{algorithm}[t]
\small
\caption{\method compression for one document.}
\label{alg:adamerge}
\begin{algorithmic}[1]
\REQUIRE patches $\mathbf{P}$, attentions $\boldsymbol{\alpha}$,
  threshold $k$, floor $c_{\min}$.
\STATE $\tau\gets\mu_{\alpha}+k\sigma_{\alpha}$;\;
       $\mathbf{P}'\!\gets\!\{\mathbf{p}_i\!:\!\alpha_i>\tau\}$ \hfill // Prune
\STATE $\mathbf{Z}\gets\textsc{WardLinkage}(\mathbf{P}')$;\;
       extract merge-cosine sequence $\mathbf{c}$.
\STATE $S\gets\{t\!:\!c_t\!\ge\!c_{\min}\}$;\;
       $t^{\star}\gets\arg\max_{t\in S}(c_t-c_{t+1})$.
\STATE $\{\mathcal{C}_k\}\!\gets\!\textsc{CutDendrogram}(\mathbf{Z},t^{\star})$.
\FOR{each cluster $\mathcal{C}_k$}
   \STATE $\widetilde{\mathbf{p}}_k\!\gets\!
   \sum_{j\in\mathcal{C}_k}\alpha_j\mathbf{p}_j\big/
   \sum_{j\in\mathcal{C}_k}\alpha_j$ \hfill // Attn-weighted centroid
\ENDFOR
\RETURN $\widetilde{\mathbf{P}}=\{\widetilde{\mathbf{p}}_k\}$.
\end{algorithmic}
\end{algorithm}

\section{Experiments}
\label{sec:exp}

\subsection{Setup}
\textbf{Benchmarks.} We evaluate on two ViDoRe suites. \emph{ViDoRe-V1} (10
short-document QA tasks: arXiv, DocVQA, InfoVQA, ShiftProject, TabFQuAD, TATDQA,
and four synthetic DocQA splits) consists of single-page documents.
\emph{ViDoRe-V2}~\cite{mace2025vidorev2} (4 long-document retrieval tasks: ESG
reports, biomedical lectures, economics reports, and human-labeled ESG reports)
consists of long, content-rich documents.
\textbf{Backbones.} ColQwen2.5-v0.2~\cite{faysse2024colpali} and
ColNomic-3B~\cite{nomicai2025}.
\textbf{Baselines.} \textsc{Identity} (no compression),
\textsc{DocPruner}~\cite{yan2025docpruner} (pruning only),
\textsc{Sem-Cluster}~\cite{ma2025light} (merging only), and
\textsc{PtM}~\cite{yan2026ptm} (state of the art). For every method with a
hyperparameter we sweep it and report the Pareto-optimal frontier; PtM's budget
$m$ is tuned over $\{1.2,1.5,2,3,4,6\}$. Concretely, each $(k,\cdot)$ configuration yields one
(compression,~nDCG@5) point per dataset, and sweeping the shared pruning
factor $k$ traces the compression axis; scores are averaged over a
benchmark's datasets and each curve shows the Pareto-optimal envelope of its
method's points. The PtM envelope additionally selects the best budget $m$
separately for every dataset, while the \method curve uses the same global
$c_{\min}$ throughout.
\method is reported at $c_{\min}\!\in\!\{0.7,0.8\}$.
The floor is deliberately coarse: detected cliffs concentrate well above it
(per-dataset medians in $[0.77,0.85]$; \Cref{sec:exp-cliff}), so its exact
value is not critical. We report results at both values rather than selecting
a single tuned one; all conclusions hold at either setting.
\textbf{Metric.} nDCG@5, averaged over the datasets within each benchmark.

\subsection{Main Results: Long Documents (ViDoRe-V2)}
\label{sec:exp-main}
\begin{figure*}[t]
  \centering
  \includegraphics[width=\textwidth]{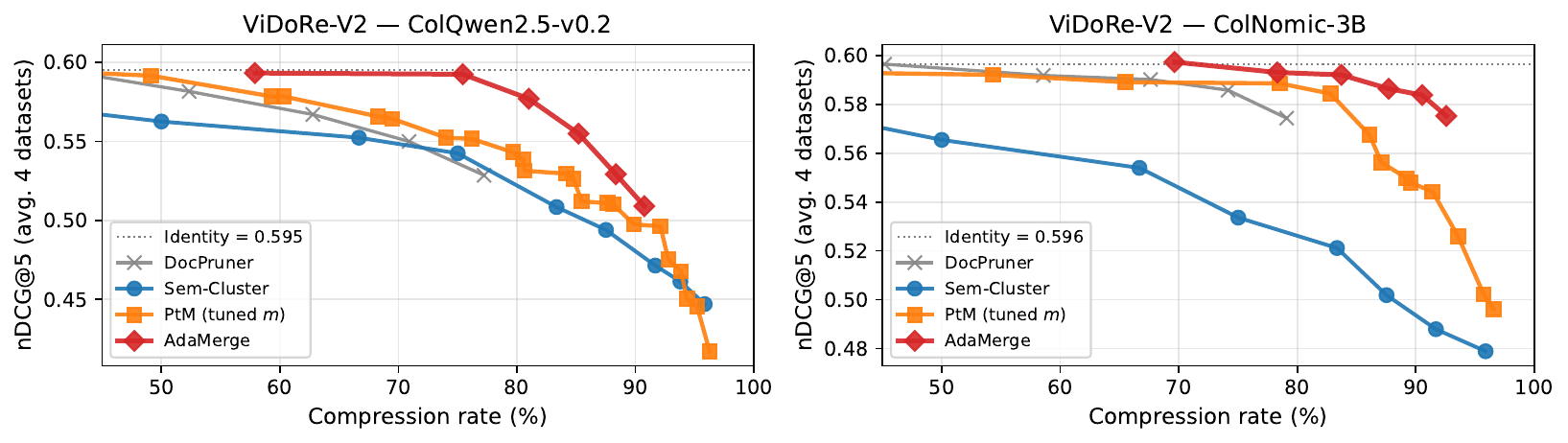}
  \caption{Compression--accuracy frontier on \textbf{ViDoRe-V2} (long documents),
  averaged over 4 datasets, for ColQwen2.5 (left) and ColNomic-3B (right).
  \method (red) uses a single global safety floor shared across all datasets;
  PtM (orange) is shown at its best per-dataset budget $m$. \method stays within
  $0.002$ of Identity up to ${\sim}75\%$ compression and leads tuned PtM across
  the operating range on both backbones.}
  \label{fig:v2}
\end{figure*}

\Cref{fig:v2} shows the compression--accuracy frontier on ViDoRe-V2 for both
backbones. \method (red) dominates the frontier throughout the operating range.
On ColQwen2.5, at $75\%$ compression \method retains $0.593$ nDCG@5 (within
$0.002$ of the uncompressed Identity at $0.595$), while tuned PtM reaches only
$0.552$; the margin persists to high compression ($+0.012$ at $90\%$). The same
pattern holds on ColNomic-3B, where \method leads tuned PtM by up to $+0.036$ in
the $85$--$90\%$ range. Pruning-only DocPruner cannot reach high compression
without collapse, and merge-only Sem-Cluster trails both pruning-aware methods.

We verify these gains are not artifacts of frontier construction with a
per-query paired Wilcoxon signed-rank test. Pooling all queries from the four
ViDoRe-V2 datasets across both backbones ($n{=}1344$), \Cref{tab:significance}
compares \method against PtM at matched pruning settings: each pair fixes the
same pruning factor $k$, which dominates the resulting compression rate
(cf.\ \Cref{tab:ablation}, where fixed-budget and adaptive variants all operate
at ${\sim}81\%$ compression), so the pairs differ essentially only in the merge
stage. \method improves per-query nDCG@5 by $+0.012$ to $+0.038$, significant
at $p<10^{-4}$ in every comparison.

\begin{table}[t]
\small
\centering
\caption{Per-query paired Wilcoxon test on ViDoRe-V2 ($n{=}1344$ queries,
both backbones pooled). All differences favor \method at $p<10^{-4}$.}
\label{tab:significance}
\begin{tabular}{cccccc}
\toprule
 & & \multicolumn{2}{c}{nDCG@5} & & \\
\cmidrule(lr){3-4}
$c_{\min}$ & $m$ & \method & PtM & $\Delta$ & $p$-value\\
\midrule
0.70 & 2 & 0.301 & 0.289 & $+0.012$ & $<10^{-4}$ \\
0.70 & 4 & 0.301 & 0.269 & $+0.032$ & $<10^{-4}$ \\
0.80 & 2 & 0.307 & 0.289 & $+0.017$ & $<10^{-4}$ \\
0.80 & 4 & 0.307 & 0.269 & $+0.038$ & $<10^{-4}$ \\
\bottomrule
\end{tabular}
\end{table}

\subsection{Short Documents (ViDoRe-V1)}
\label{sec:exp-v1}
\begin{figure*}[t]
  \centering
  \includegraphics[width=\textwidth]{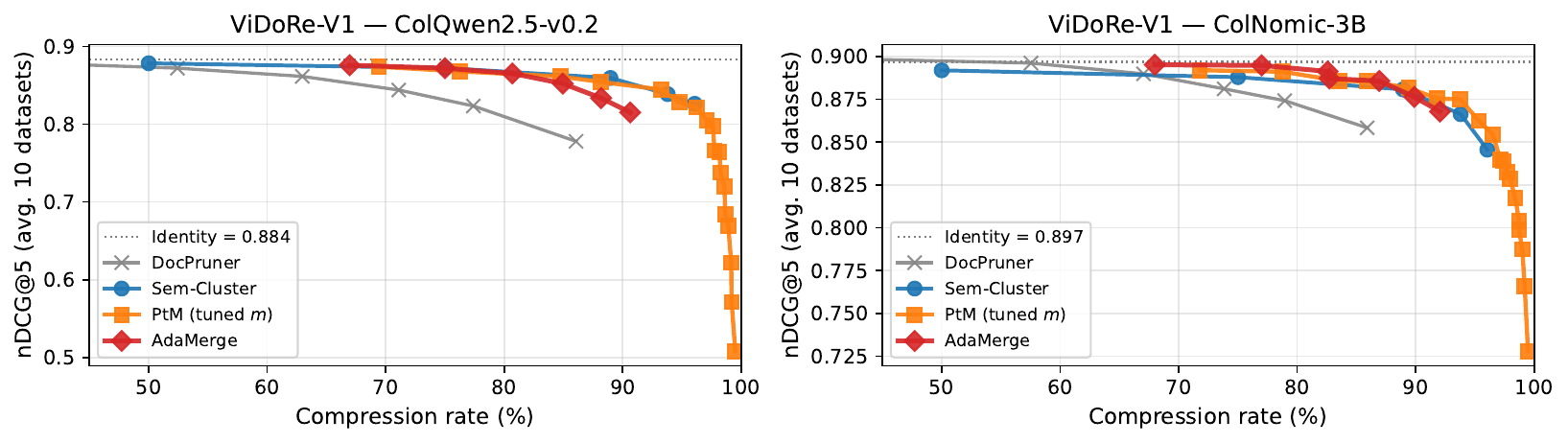}
  \caption{ViDoRe-V1 (short documents), averaged over 10 datasets, for
  ColQwen2.5 (left) and ColNomic-3B (right). All merging methods are
  near-lossless and statistically indistinguishable; \method matches tuned PtM
  with no per-dataset tuning on both backbones.}
  \label{fig:v1}
\end{figure*}

\Cref{fig:v1} repeats the comparison on ViDoRe-V1 for both backbones. Here the
picture is different: every merging method (Sem-Cluster, PtM, and \method) sits
close to the uncompressed Identity across the whole range and the curves are
nearly indistinguishable (within ${\sim}0.01$ nDCG@5 up to $85\%$ compression on
both ColQwen2.5 and ColNomic-3B). This is expected: ViDoRe-V1 documents are
single pages with little intra-page redundancy, so few merges are needed and
\emph{where} to stop matters little; compression is ``easy'' and offers little
room to separate methods. \method matches tuned PtM in this regime \emph{without}
the per-dataset budget search PtM requires. The advantage of adaptive cutting
therefore emerges precisely where it should: on long, redundant documents
(\Cref{fig:v2}), where many merges are possible and the stopping point is
consequential.

\subsection{The Cliff is Concentrated across Documents}
\label{sec:exp-cliff}
\begin{figure}[t]
  \centering
  \includegraphics[width=\columnwidth]{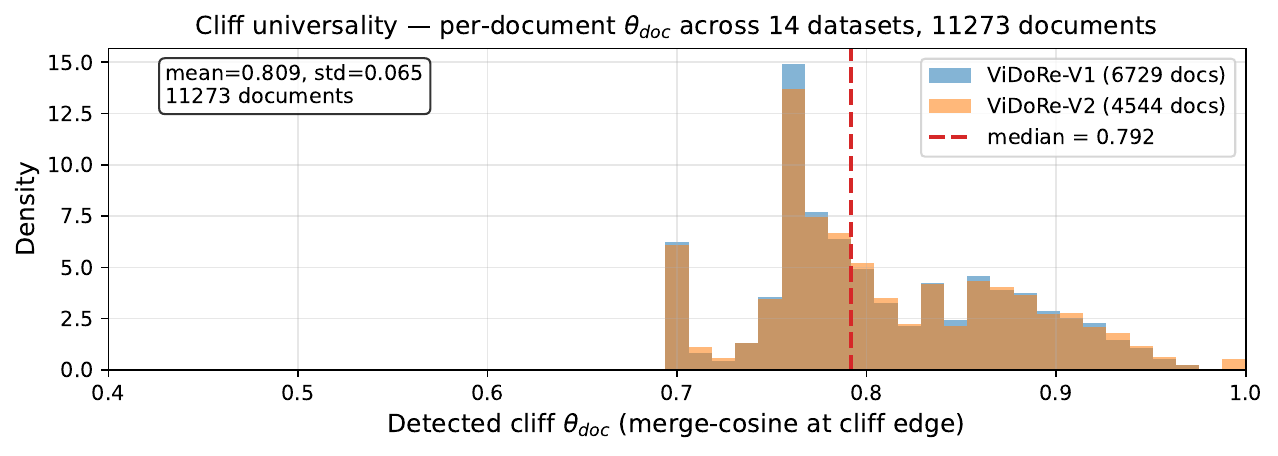}
  \caption{Distribution of the per-document cut location $\theta_{\text{doc}}$
  over $11{,}273$ documents (14 datasets, ColQwen2.5). Pooled median
  ${\approx}0.79$; per-dataset medians span $[0.77,0.85]$.}
  \label{fig:cliff}
\end{figure}

\Cref{fig:cliff} reports the distribution of the per-document cut location
$\theta_{\text{doc}}\!=\!c_{t^{\star}}$ over the $11{,}273$ documents of all 14
datasets (ColQwen2.5). The distribution is tight, with a pooled median of
$0.79$ and standard deviation $0.065$, and per-dataset medians spanning only
$[0.77,0.85]$ despite the heterogeneity of the corpora (arXiv papers, ESG
reports, medical lectures, tabular figures).\footnote{$\theta_{\text{doc}}$ is
bounded below by the safety floor $c_{\min}$, so the left tail is truncated by
construction; the concentration we rely on is the agreement of the
\emph{per-dataset medians}, which lie well above the floor.} This concentration
indicates that the merge boundary is a document-level property recoverable from
$\mathbf{c}$, supporting per-document detection over a fixed global budget.

\subsection{Component Ablation and Efficiency}
\label{sec:exp-ablation}
\method makes two changes over PtM: (a)~the per-document adaptive cut replaces
the fixed budget, and (b)~the attention-weighted centroid replaces the
unweighted mean. \Cref{tab:ablation} disentangles them on ViDoRe-V2
(ColQwen2.5, averaged over 4 datasets) at a representative operating point
(${\sim}81\%$ compression). The adaptive cut is the dominant contributor
($+0.039$ nDCG@5 over PtM), and attention weighting adds a further $+0.009$ on
top, confirming both components contribute and the cut is primary.

\begin{table}[t]
\small
\centering
\caption{Component ablation on ViDoRe-V2 (ColQwen2.5, 4 datasets) at
${\sim}81\%$ compression. The adaptive cut drives the gain; attention weighting
adds a smaller consistent improvement.}
\label{tab:ablation}
\begin{tabular}{llcc}
\toprule
Variant & Cut & Centroid & nDCG@5 \\
\midrule
PtM (baseline)            & fixed $m$          & mean          & 0.530 \\
PtM-attn                  & fixed $m$          & attn          & 0.533 \\
AdaMerge-mean (ablation)  & adaptive           & mean          & 0.568 \\
\textbf{\method} (full)   & \textbf{adaptive}  & \textbf{attn} & \textbf{0.577} \\
\bottomrule
\end{tabular}
\end{table}

\textbf{Efficiency.} On ViDoRe-V2 with ColQwen2.5, \method compresses a document
in ${\approx}9$--$10$~ms on a single CPU thread, versus ${\approx}8$~ms for PtM;
DocPruner is faster (${\sim}0.3$~ms) but cannot reach the same operating points.
The gap is inherent: PtM can halt the agglomeration once $\lceil N'/m\rceil$
clusters remain, whereas \method computes the full linkage to locate the cliff,
adding ${\approx}1$--$2$~ms per document. This one-time indexing cost is offset
at deployment by eliminating PtM's per-dataset grid search over $m$.

\section{Conclusion}
\label{sec:concl}
We revisited where patch merging should stop in multi-vector VDR compression and
identified a sharp, consistently located cliff in each document's merge-cosine
sequence. Building on it, \method detects the cliff per document and forms
attention-weighted centroids, replacing PtM's per-dataset cluster budget with a
single global constant. \method significantly outperforms tuned PtM on
long-document ViDoRe-V2 ($p<10^{-4}$) and matches it on short-document
ViDoRe-V1 with no per-dataset tuning, at ${\approx}10$~ms per document. Future
work includes characterizing the cliff theoretically and extending per-document
adaptive compression to non-ColPali multi-vector encoders.

\section*{GenAI Usage Disclosure}
The core research contributions of this paper, namely the observation of the
merge-cosine cliff, the design of the \method method (per-document adaptive
cutting and attention-weighted centroids), the experimental protocol, and the
interpretation of all results, were conceived and carried out by the authors.
Generative AI tools were used only in a supporting capacity: (i)~assisting in
writing and refactoring the experiment, evaluation, and plotting code;
(ii)~helping to orchestrate large-scale experiment runs and produce figures from
author-specified results; and (iii)~light editing and polishing of the
manuscript text (grammar, phrasing, and \LaTeX{} formatting of author-written
content). No part of the paper's technical claims, methodology, or experimental
results was generated solely by a generative model. The authors verified all
reported numbers, figures, and statements, and take full responsibility for the
content of this paper.

\bibliographystyle{ACM-Reference-Format}
\bibliography{references}

\end{document}